\documentclass[letterpaper]{article} % DO NOT CHANGE THIS
\usepackage{aaai24}  % DO NOT CHANGE THIS
\usepackage{times}  % DO NOT CHANGE THIS
\usepackage{helvet}  % DO NOT CHANGE THIS
\usepackage{courier}  % DO NOT CHANGE THIS
\usepackage[hyphens]{url}  % DO NOT CHANGE THIS
\usepackage{graphicx} % DO NOT CHANGE THIS
\usepackage{natbib}  % DO NOT CHANGE THIS AND DO NOT ADD ANY OPTIONS TO IT
\usepackage{caption} % DO NOT CHANGE THIS AND DO NOT ADD ANY OPTIONS TO IT
\usepackage{algorithm}
\usepackage{algorithmic}

\usepackage{amsmath}
\usepackage{amsmath,amssymb}
\usepackage{bm}
\usepackage{dsfont}

\usepackage[dvipsnames]{xcolor}
\usepackage{breqn}
\usepackage{booktabs}
\usepackage{multirow}

\usepackage{newfloat}
\usepackage{listings}
\DeclareCaptionStyle{ruled}{labelfont=normalfont,labelsep=colon,strut=off} % DO NOT CHANGE THIS
\floatstyle{ruled}
\newfloat{listing}{tb}{lst}{}
\floatname{listing}{Listing}
\title{ReAGent: A Model-agnostic Feature Attribution Method for Generative Language Models}
\author{
    Zhixue Zhao, Boxuan Shan
}
\affiliations{
    Department of Computer Science, University of Sheffield \\
    \{zhixue.zhao, bshan3\}@sheffield.ac.uk
}

\usepackage{bibentry}
\begin{document}

\maketitle

\begin{abstract}
Feature attribution methods (FAs), such as gradients and attention, are widely employed approaches to derive the importance of all input features to the model predictions.
Existing work in natural language processing has mostly focused on developing and testing FAs for encoder-only language models (LMs) in classification tasks. However, it is unknown if it is faithful to use these FAs for decoder-only models on text generation, due to the inherent differences between model architectures and task settings respectively. Moreover, previous work has demonstrated that there is no `one-wins-all' FA across models and tasks. This makes the selection of a FA computationally expensive for large LMs since input importance derivation often requires multiple forward and backward passes including gradient computations that might be prohibitive even with access to large compute. 
To address these issues, we present a model-agnostic FA for generative LMs called Recursive Attribution Generator (ReAGent). Our method updates the token importance distribution in a recursive manner. For each update, we compute the difference in the probability distribution over the vocabulary for predicting the next token between using the original input and using a modified version where a part of the input is replaced with RoBERTa predictions. Our intuition is that replacing an important token in the context should have resulted in a larger change in the model's confidence in predicting the token than replacing an unimportant token.
Our method can be universally applied to any generative LM without accessing internal model weights or additional training and fine-tuning, as most other FAs require. We extensively compare the faithfulness of ReAGent with seven popular FAs across six decoder-only LMs of various sizes. The results show that our method consistently provides more faithful token importance distributions.\footnote{Our code: \url{https://github.com/casszhao/ReAGent}}
\end{abstract}

\section{Introduction}

Feature attribution (FA) techniques are popular post-hoc methods to provide explanations for model predictions. FAs generate token-level importance scores to highlight the contribution of each token to a prediction~\citep{denil2014extraction, li2015visualizing, jain-etal-2020-learning, kersten-etal-2021-attention}. The top-k important tokens are typically considered as the prediction rationale~\citep{zaidan-etal-2007-using, kindermans2016investigating, sundararajan2017axiomatic, deyoung-etal-2020-eraser}. The quality of a rationale is often evaluated using faithfulness metrics which measure to what extent the rationale accurately reflects the decision mechanisms of the model~\cite{deyoung-etal-2020-eraser, jacovi-goldberg-2020-towards, ding-koehn-2021-evaluating,zhao-aletras-2023-incorporating}.\footnote{Plausibility is another aspect of rationale quality, measuring how understandable it is to humans~\citep{jacovi-goldberg-2020-towards} and it is out of the scope of our study.} 

\begin{figure}[!t] 
\centering
    \includegraphics[trim={1cm 0cm 1cm 1cm}, width=.98\columnwidth]{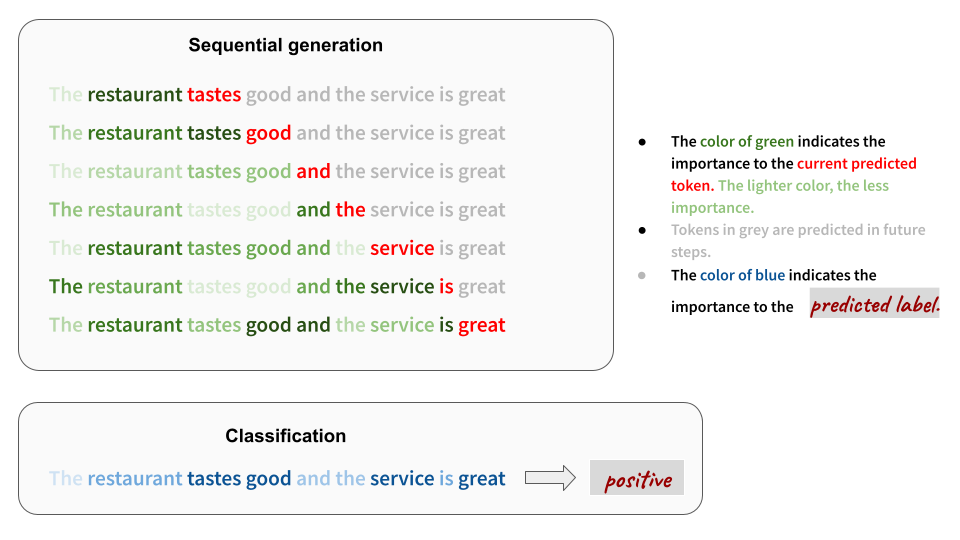}\\
    \caption{Input importance distributions for a generative task (top) and a classification task (bottom) using a toy FA\@.
    }\label{fig:teaser2}
\end{figure}

The faithfulness of FAs is well-studied in the context of text classification tasks, particularly with encoder-only language models (LMs), e.g.\ BERT-based models~\cite{ding-koehn-2021-evaluating, zhou-etal-2022-exsum, zhao-etal-2022-impact, attanasio-etal-2023-ferret}. A range of FAs has been developed and evaluated in such a context, e.g.\ propagation-based methods that rely on forward-backward passes to the model for computing importance~\citep{kindermans2016investigating, sundararajan2017axiomatic, atanasova-etal-2020-diagnostic}, attention-based methods that use attention weights as an indication of importance~\citep{serrano-smith-2019-attention,jain-etal-2020-learning,abnar-zuidema-2020-quantifying}, and occlusion-based methods that probe the model behavior with corrupted input~\cite{lei-etal-2016-rationalizing,bastings-etal-2019-interpretable, bashier-etal-2020-rancc}. % Surrogate-model based

However, it is non-trivial and not guaranteed to be faithful to apply these FAs on decoder-only LMs for text generation directly, due to the inherent differences between encoder and decoder-only models and task settings respectively, e.g.\ different attention mechanisms. Moreover, FAs should produce an attribution distribution per token position for decoder-only LMs in generation tasks, while only a single attribution distribution needs to be computed for fine-tuned encoder-only models. 
Figure~\ref{fig:teaser2} shows the difference in complexity of applying a toy FA on a decoder and encoder-only LM respectively. Previous work has demonstrated that there is no `one-wins-all' FA across models and tasks \citep{atanasova-etal-2020-diagnostic}.
Therefore, a separate comparison of various FAs is required for each task and model. However, this is computationally expensive for large LMs (e.g.\ some FAs may additionally require forward-backward passes). In some cases, FAs might not even be applicable to black-box models that are available only through an API that might provide access to the predictive likelihood but not the internal weights, e.g.\ the OpenAI API.\footnote{\url{https://platform.openai.com/docs/api-reference/introduction}}

\citet{vafa-etal-2021-rationales} proposed using a greedy search to find the shortest subset of the input as the rationale for generation tasks. However, their method only highlights important tokens in a binary manner, i.e.\ important or not. More importantly, this method requires model fine-tuning which is prohibitive for large LMs and it might not truly reflect the original models' inner reasoning process but the new fine-tuned one. 

To address these challenges, we propose Recursive Attribution Generator (ReAGent), a FA for decoder-only LMs that computes the input importance distribution for each token position recursively by replacing tokens from the original input with tokens predicted by RoBERTa where the rest are masked. Our intuition is that replacing important tokens should result in larger predictive likelihood differences. We make the following contributions:

\begin{itemize}

    \item ReAGent does not require accessing the model internally and can be easily applied to any generative LM;\@
    \item We empirically show that ReAGent is consistently more faithful than seven popular FAs by conducting a comprehensive suite of experiments, covering three tasks and six LMs of varying sizes from two different model families;
    \item Our method does not need to retain any gradients for computing input importance but only relies on forward passes~\citep{bach2015pixel,lundberg2017unified,selvaraju2017grad, ancona2018towards}. This minimizes the memory footprint and compute required.
    
\end{itemize}

\section{Related Work}

\subsection{Post-hoc FAs}

Post-hoc explanation methods such as FA techniques are applied retrospectively by seeking to extract explanations after the model makes a prediction.
Most FAs have been proposed in the context of classification tasks, where a sequence input \(x_{1}, \ldots, x_{t-1}\) is associated with a true label \(y\) and a predicted label \(\hat{y}\). The underlying goal is to identify which parts of the input contribute more toward the prediction \(\hat{y}\)~\citep{lei-etal-2016-rationalizing}. Figure~\ref{fig:teaser2} (bottom) shows an example. 
Most FAs generally fall into propagation, attention, and occlusion-based categories.

Propagation-based (i.e.\ gradient-based) methods derive the importance for each token by computing gradients with respect to the input~\citep{denil2014extraction}. Building upon this, the input token vector is multiplied by the gradient (Input x Gradient)~\citep{denil2014extraction}, while Integrated Gradients compares the input with a null baseline input when computing the gradients with respect to the input~\citep{sundararajan2017axiomatic}. Other propagation-based FAs include Layer-wise Relevance Propagation~\citep{bach2015pixel}, GradientSHAP~\citep{lundberg2017unified}, and DeepLIFT~\citep{selvaraju2017grad, ancona2018towards}.  \citet{9810053} offer a comprehensive overview of different propagation-based FAs.

Attention-based FAs are applied to models that include an attention mechanism to weigh the input tokens. The assumption is that the attention weights represent the importance of each token. These FAs include scaling the attention weights by their gradients, taking the attention scores from the last layer, and recursively computing the attention in each layer~\citep{serrano-smith-2019-attention,jain-etal-2020-learning,abnar-zuidema-2020-quantifying}. 

Occlusion-based methods measure the difference in model prediction between using the original input and a corrupted version of the input by gradually removing tokens~\cite{lei-etal-2016-rationalizing, nguyen-2018-comparing,bastings-etal-2019-interpretable, bashier-etal-2020-rancc} 
The underlying idea is that removing important tokens will lead the model to flip its prediction or a significant drop in the prediction confidence.

Another line of work has focused on learning feature attributions with a modified model or a separate explainer model~\cite{ribeiro2016should, lundberg2017unified, kindermans2019reliability, bashier-etal-2020-rancc, kokhlikyan2021investigating, hase2021out}. For example, LIME trains a linear classifier that approximates the model behavior in the neighborhood of a given input. SHAP (SHapley Additive exPlanations) learns the predictive probability of using different combinations of tokens and then computes the marginal contribution of each token towards the target prediction~\citep{lundberg2017unified}.

The methods above have specific model dependencies (e.g.\ attention) or require specific training or fine-tuning, which makes it difficult to apply them to decoder-only models for generation tasks directly.

%%%%%%%%%%%%%%%%%%%%5
\subsection{FAs for Generative Models}
In text generation tasks, the model (often decoder-only) needs \(n\) forward passes to generate a sequence of \(n\) tokens. As shown in Figure~\ref{fig:teaser2}, a different token importance distribution is computed for each predicted token. 

\citet{vafa-etal-2021-rationales} propose a greedy search approach to find a minimum subset of the input that maximizes the probability of the target token which is considered as the rationale for generative LMs.  
Their method identifies if a token belongs to the rationale or not, i.e.\ binary prediction, rather than an importance distribution. This makes it hard to evaluate if such a rationale is more faithful than others. Furthermore, this method requires fine-tuning the original model to support blank positions (i.e.\ removed tokens) which is rather challenging and computationally inefficient for large LMs.  
\citet{cifka-liutkus-2023-black} estimate the importance of each token by exploring how different lengths of their left-hand side context impact the predictive probability of the target token. They describe the estimated importance output as ``differential importance scores'', and point out that a higher score should not be interpreted as the token is more important for the prediction, but it should be considered as how much ``new information'' the token adds to the context that the other tokens have not covered yet. 

To the best of our knowledge, no previous work has proposed a model-agnostic FA for generative LMs.

%%%%%%%%%%%%%%%%%%  Preliminaries
\section{Preliminaries}

\subsection{Generative Language Modeling}

In generative language modeling, the input is a sequence of tokens, \(\bm{X}=\left[x_{1}, \ldots, x_{t-1}\right]\), i.e.\ the initial input (often called a prompt). 
The goal is to learn a model \(f_{\theta }\) that approximates the probability \(p_{\theta }\) of the token sequence \(x_{1}, \ldots, x_{t-1}\). Here, we assume that \(f_{\theta}\) is a specific pre-trained generative LM, such as GPT \citep{brown2020language}, for predicting the probability of the next token \(x_t\) conditioned to the context \(x_{1}, \ldots, x_{t-1}\) and parameterized by \(\theta \): 
\begin{equation}
\scriptsize
p_{\theta}\left(x_{1}, \ldots, x_{t-1}\right)=f_{\theta}\left(x_{1}\right) \prod_{t=2}^{T} f_{\theta}\left(x_{t} \mid x_{1}, \ldots, x_{t-1}\right)
\end{equation}

\subsection{Input Importance for Generative LMs}

Assuming an underlying model \(f_{\theta}\), we want to compute the input importance distribution for each predicted token \(\hat{x} \subset X\) given the input tokens \(X=x_{1}, \ldots, x_{t-1}\).

A FA method \(e_t\) applied on position \(t\) produces an importance distribution \(S_t=s_{1}, \ldots, s_{t-1}\) for a target token \(x_t\): 
\begin{equation}
\small e_t(f, \theta, x_{1}, \ldots, x_{t-1}, x_t) \rightarrow S_t, t\in \{1,\ldots,n\}
\end{equation}
We assume that a higher \(s_i\) denotes higher importance. \(e\) is applied \(n\) times for generating importance distributions for \(n\) tokens. For a model agnostic \(e_t\), access to model parameters \(\theta \) is not required.

%%%%%%%%%%%%%%%%%%%%%%%%%%%%%%% METHOD
\section{Recursive Attribution Generator (ReAGent)}

%%%%%%%%%%%%%%%%%%%%%%%%%%%%%%  METHOD 

Our aim is to compute the input importance distribution per predicted token by being (1) model-agnostic (i.e.\ we should not rely on an attention mechanism); and (2) able to support black-box LMs with no access to their weights and architecture details (i.e.\ we should not rely on backward passes for computing gradients). 

Inspired by occlusion-based FA methods~\citep{zeiler2014visualizing,nguyen-2018-comparing}, we assume that replacing more important tokens in the input context \(x_{1}, \ldots, x_{t-1}\) will result in larger drops in the predictive likelihood of the target token \(x_t\). Therefore, we propose ReAGent, a new method that iteratively replaces tokens from the context \(x_{1}, \ldots, x_{t-1}\) to compute token importance.

The overall process (shown in Algorithm~\ref{alg:pseduo_code}) consists of: (1) initializing importance scores randomly (Step 1); (2) starting an iteration loop for \textit{updating importance scores} (Steps 3 to 6) till meeting the \textit{stopping condition} (Step 2).

\begin{algorithm}%[tb]
\begin{flushleft}
    \small
    \caption{Recursive Attribution Generator}\label{alg:pseduo_code}
    \textbf{Input}: LM \(f\), context \(x_{1}, \ldots, x_{t-1}\), target token \(x_{t}\) \\
    \textbf{Output}: \(\textbf{S}_t = \{{s_{1},\dots,s_{t-1}}\} \)
    \begin{algorithmic}[1] %[1] enables line numbers
        \STATE{Randomly initialize importance scores \(\bm{S}_t\)}
        \WHILE{!StoppingCondition~(\(\bm{S}_t\), \(x_{t}\))}
                \STATE{\(\mathcal{R}\) \(\gets \) randomly select tokens \(\mathcal{R} \in x_{1}, \ldots, x_{t-1}\)} \\
                \STATE{\(\hat{x}_{1}, \ldots, \hat{x}_{t-1}\) \(\gets \) replace \(\mathcal{R}\) on \(x_{1}, \ldots, x_{t-1}\) with tokens predicted by RoBERTa} \\
                \STATE{\(\Delta p\) \(\gets \) \(p(x_{t}|x_{1}, \ldots, x_{t-1}) - p(x_{t}|\hat{x}_{1\dots t-1})\)} \\
                \STATE{update importance scores \(s_{1},\dots, s_{t-1}\) by \(\Delta p\) and \(\mathcal{R}\)} \\
        \ENDWHILE{}
        \STATE{\textbf{return} \(\bm{S}_t\)}
    \end{algorithmic}
\end{flushleft}
\end{algorithm}

%%%%%%%%%%%%%%%%%%%%%%%%%%%%%%  METHOD UPDATING
\subsection{Computing Importance Scores}

To compute the importance of the context tokens \(x_{1}, \ldots, x_{t-1}\), we repeat the following steps until a stopping condition is met.

\paragraph{Step 3: Context tokens to be replaced} 
We first randomly choose a set of tokens \(\mathcal{R}\) (\(r\% \) of the entire sequence, \(r=30\)) from \(x_{1}, \ldots, x_{t-1}\) to be replaced. 

\paragraph{Step 4: Replacement tokens}
A replacing function \(\bm{C}\) replaces each token in \(\bm{X}\) with the token predicted by RoBERTa.\footnote{We also test another two methods providing tokens to replace. However, they empirically perform poorly. We include the results of these two methods in Section~\ref{app:alternative_find_token_replace} in the Appendix.} A new sequence is generated for the occlusion of \(\bm{X}\) later. 
\begin{equation}
\small
\begin{split}
\bm{C}(\bm{X}) =& [ \hat{x}_1, \hat{x}_2, \dots \hat{x}_{t-1} ] \\ & \quad \sim U(\mathcal{X}^{(g)}([x_1, x_2, \dots, x_{t-1}]))
\end{split}
\end{equation}

\paragraph{Steps 5 \& 6: Updating importance scores}
The importance distribution \(\bm{S}_t\) is computed based on the selection of tokens, \(\mathcal{R}\), and the predictive difference, \(\Delta p\) in Eq.~\ref{alg:delta_p}, after \(\mathcal{R}\) replaced by \(\bm{C}(\bm{X})\) as following:
{\small
\begin{align}
    p^{(o)}_{t} &= p(x_t|\bm{X}) \label{alg:def_p_o}\\
    p^{(r)}_{t} &= p(x_t|(\bm{M}(\bm{X}, \overline{\mathcal{R}}) + \bm{M}(\bm{C}(\bm{X}), \mathcal{R}))) \label{alg:def_p_r}\\
    \Delta p_{t} &= p^{(o)}_{t} - p^{(r)}_{t} \label{alg:delta_p}\\
    \Delta \bm{S}_t &= \bm{M}(\Delta p_{t} \cdot \mathds{1}^{|\bm{X}|}, \mathcal{R}) + \bm{M}(-\Delta p_{t} \cdot \mathds{1}^{|\bm{X}|}, \overline{\mathcal{R}}) \label{alg:delta_s} \\ % need to change from here
    \bm{S}^{(l)}_t &= \bm{S}^{(l)}_{n-1} + \text{logit}\left(\frac{\Delta \bm{S}_t + 1}{2}\right)\label{alg:update_logit_s} \\
    \bm{S}_t &= \text{softmax}(\bm{S}^{(l)}_t)\label{alg:update_s}
\end{align}
}

\noindent where \(p^{(o)}_{t}\) and \(p^{(r)}_{t}\) are predictive probabilities for target \(x_{t}\) when using the original sequence and the sequence with replaced tokens, respectively. 
\(\bm{M}(\bm{X}, \mathcal{R})\) is a masking function that results in token replacements only for \(x_i\) selected in step 3. 
Finally, \(\bm{S}_t\) is obtained by adding \(\Delta \bm{S}_t\) with \(\bm{S}_{t-1}\) in logistic scale.

%%%%%%%%%%%%%%%%%%%%%%%%%%%%%%  METHOD STOPPING
\subsection{Step 2: Stopping Condition} 

A stopping condition is set up for the iterative updating process. When prompting the model with top-\(n\) (default \(70\% \)) unimportant tokens replaced with RoBERTa predictions, if the top-k (default set to 3) predicted candidate tokens (by the model to explain) contain the target token \(x_{t}\), the iterative process stops.

%%%%%%%%%%%%%%%%%%%%%% EXPERIMENT SETUP %%%%%%%%%%%%%%%%%%%%%%%%%

\section{Experimental Setup}\label{sec:experimental_setup}

%%% TASK
\subsection{Datasets}

\begin{table}[ht]
\resizebox{\columnwidth}{!}{%
\begin{tabular}{@{}lccl@{}}
\toprule
{\bf Dataset} & {\bf Length} & {\bf \#Data} & {\bf Prompt Example} \\ \midrule
LongRA & 36 & 37--149 & \parbox{9cm}{``When my flight landed in \underline{Japan}, I converted my currency and slowly fell asleep. (I had a terrifying dream about my grandmother, but that's a story for another time). I was staying in the capital, \underline{\qquad\qquad}''} \\ \midrule

TellMeWhy & 50 & 200 & \parbox{9cm}{``Joe ripped his backpack. He needed a new one. He went to Office Depot. They had only one in stock. Joe was able to nab it just in time.Why did He need a new one?''} \\ \midrule

WikiBio & 35 & 238 & \parbox{9cm}{``Rudy Fernandez (1941–2008) was a labor leader and civil rights activist from the United States. He was born in San Antonio, Texas, and was the son of Mexican immigrants.''} \\ \bottomrule
\end{tabular}%
}
\caption{Summary of Tasks. Note that different models have different LongRA data as we only keep ones that have the same predictions with and without distractors (sentences in parenthesis). Length refers to the average token count of input sequences.}\label{tab:data_summary}
\end{table}

We experiment with three generic text generation datasets for prompting LMs with different contextual levels, following recent related work~\citep{vafa-etal-2021-rationales,lal-etal-2021-tellmewhy,gehrmann2021gem, manakul2023selfcheckgpt}.

\begin{itemize}
    \item \textbf{Long-Range Agreement (LongRA)} is designed by~\citet{vafa-etal-2021-rationales} to allure the model to predict a specific word. The dataset uses template sentences based on word pairs from~\citet{mikolov2013efficient} such that the two words, e.g. \textit{Japan} and \textit{\underline{Tokyo}} in Table~\ref{tab:data_summary}, are semantically or syntactically-related. The input includes one word from a pair and aims to prompt the model to predict the other word in the pair.
    A distractor sentence that contains no pertinent information about the word pair is also inserted, e.g.\ in () in Table~\ref{tab:data_summary}. Following~\citet{vafa-etal-2021-rationales}, we only keep examples where the prediction is the same both with and without the distractor.
    \item \textbf{TellMeWhy} is a generation task for answering why-questions in narratives. The input contains a short narrative and a question concerning why characters in a narrative perform certain actions~\cite{lal-etal-2021-tellmewhy}. We randomly sample 200 examples.
    \item \textbf{WikiBio} is a dataset of Wikipedia biographies. We take the first two sentences as a prompt similar to~\citet{manakul2023selfcheckgpt}. The model is expected to continue generating the biography. This task is intuitively more open than the two tasks above.
\end{itemize}

%%% MODEL

\subsection{Models}

We test six large LMs, three models from the GPT~\cite{radford2019language, brown2020language, mesh-transformer-jax, gpt-j} and three from the OPT family~\cite{zhang2022opt}. We choose our models to span from hundreds of millions to a few billions of parameters (354M, 1.5B and 6B parameters for GPT;\@ and 350M, 1.3B, and 6.7B parameters for OPT), as we want to explore how the model size affects the faithfulness of each FA\@. All models used are publicly available.~\footnote{We use checkpoints from the Huggingface library for each model with the following identifiers: \texttt{gpt2- medium}, \texttt{gpt2-xl}, \texttt{EleutherAI/gpt-j-6b}, \texttt{facebook/opt-350m}, \texttt{facebook/opt-1.3b}, \texttt{KoboldAI/OPT-6.7B-Erebus}.}

% FA
\subsection{Feature Attribution Methods}\label{sec:FAs}

For comparison, we test seven widely-used FAs~\citep{atanasova-etal-2020-diagnostic,vafa-etal-2021-rationales}: 

\begin{itemize}
    
    \item{\bf Input X Gradient:} embedding gradients multiplied by the embeddings~\cite{denil2014extraction}.
    \item{\bf Integrated Gradients:} Integrate overall gradients using a linear interpolation between a baseline input (all zero embeddings) and the original input~\citep{sundararajan2017axiomatic}.
    \item{\bf Gradient Shap:} Compute the gradient with respect to randomly selected points between the inputs and a baseline distribution~\cite{lundberg2017unified}.
    \item{\bf Attention:} It is computed averaged attentions across all layers and heads~\citep{jain-etal-2020-learning}.
    \item{\bf Last Attention:} The last-layer attention weights averaged across heads~\citep{jain-etal-2020-learning}.
    \item{\bf Attention rollout:} Recursively computing the token attention in each layer, e.g.\ computing the attention from all positions in layer \(l_i\) to all positions in layer \(l_j\), where \(j < i\)~\citep{abnar-zuidema-2020-quantifying}.
    \item{\bf LIME:} For each input, train a linear surrogate model using data points randomly sampled locally around the prediction~\citep{ribeiro2016should}.
    \item{\bf Random} Following~\citet{chrysostomou2022flexible}, we take a random attribution baseline (i.e.\ randomly assigning importance scores) as a yardstick to have a comparable analysis across data.
\end{itemize}

We exclude FAs that demand additional model modifications, e.g.\ greedy search~\citep{vafa-etal-2021-rationales}.\footnote{We provide a comparison with the greedy search on GPT2--354M in the Appendix. Due to the expensive fine-tuning, we do not apply it to other models.}

%%% EVALUATION

\subsection{Faithfulness Evaluation}
For evaluation, we use normalized Soft-Sufficiency (Soft-NS) and comprehensiveness (Soft-NC) that measure the faithfulness of the full importance distribution proposed by~\citet{zhao-aletras-2023-incorporating}. The metrics are defined as follows:
\begin{equation}\label{eq:soft_eq}
\small
\begin{aligned}
    \text{Soft-S}(\mathbf{X}, x_{t}, \mathbf{X'}) = 1 - \max(0, p(\hat{y}| \mathbf{X})- p (\hat{y}|\mathbf{X'})) \\
    \text{Soft-NS}(\mathbf{X}, \hat{y}, \mathbf{X'}) =
    \frac{\text{Soft-S}(\mathbf{X}, \hat{y}, \mathbf{X'}) - \text{S}(\mathbf{X}, \hat{y}, 0)}{1 - \text{S}(\mathbf{X}, \hat{y}, 0)} \\
    \text{Soft-C}(\mathbf{X}, \hat{y}, \mathbf{X'}) = \max(0, p(\hat{y}| \mathbf{X})- p (\hat{y}|\mathbf{X'})) \\
    \text{Soft-NC}(\mathbf{X}, \hat{y}, \mathbf{X'}) = \frac{\text{Soft-C}(\mathbf{X}, \hat{y}, \mathbf{X'})}{1 - \text{S}(\mathbf{X}, \hat{y}, 0)} \\
\end{aligned}
\end{equation}
\noindent where \(\mathbf{X'}\) is obtained by using \(q=1-s_i\) in Eq.~\ref{eq:soft-perturbation} for each token vector \(\mathbf{x'_i}\). 
Given a token vector \(\mathbf{x}_i\) from the input \(\mathbf{X}\) and its FA score \(s_i\), the input is soft-perturbed as follows:
\begin{align}
           \mathbf{x'_i} = \mathbf{x_i} \odot \mathbf{e_i},\: \mathbf{e_i}\sim \operatorname {Ber} (q)
           \label{eq:soft-perturbation}
\end{align}
\noindent where \(Ber\) a Bernoulli distribution and \(\mathbf{e}\) a binary mask vector of size \(n\). \(Ber\) is parameterized with probability \(q\): 
\begin{equation}\label{eq:bermouli}
   q = 
\begin{cases}
    s,   & \text{if retaining elements}\\
    1-s, & \text{if removing elements}
\end{cases}
\end{equation}

\noindent {\bf Adapting Soft-NS \& NC for Generative LMs}

Soft-NS and Soft-NC have been proposed for evaluating FAs in the context of encoder-only models. To evaluate FAs on generation tasks, there is one issue to address when adapting Soft-NS and Soft-NC to generative tasks and models: the lack of predictive likelihood of the predicted label. Naturally, we can treat each token as a label. However, in empirical experiments, we find that the high dimension of the output leads to high sensitivity of the predictive probability of the specific token. In other words, \(p(\hat{y}| \mathbf{X})- p (\hat{y}|\mathcal{R})\) and \(p(\hat{y}| \mathbf{X})- p (\hat{y}|\mathbf{X}_{\backslash\mathcal{R}})\) often approximate to \(p(\hat{y| \mathbf{X}})\). Re-scaling is a potential but hassle solution due to the fact that different inputs might need different scaling. 

Therefore, we propose to measure the Hellinger distance between prediction distributions over the vocabulary to assess the changes in model predictions. Essentially, we replace \(p(\hat{y}| \mathbf{X})- p (\hat{y}|\mathbf{X'})\) in Eq.~\ref{eq:soft_eq} with the Hellinger distance. 
Considering two discrete probability distributions \(\boldsymbol{P}_{X, t}=\left[p_{1, t}, \ldots, p_{v, t}\right]\) and \(\boldsymbol{P}_{X', t}=\left[p'_{1, t}, \ldots, p'_{v, t}\right]\), the Hellinger distance is formally defined as:

\begin{equation}
\small
\label{equ:hellinger}
\begin{aligned}
\Delta \boldsymbol{P}_{X', t} & = H(\boldsymbol{P}_{X, t}, \boldsymbol{P}_{X', t}) \\
& = \frac{1}{\sqrt{2}} \sqrt{\sum_{i=1}^{v}{\left(\sqrt{p_{i, t}}-\sqrt{p'_{i, t}}\right)}^{2}}
\end{aligned}
\end{equation}

\noindent where \(\boldsymbol{P}_{X, t}\) is the probability distribution over the entire vocabulary (of size \(v\)) when prompting the model with the full-text, \(X\). \(\boldsymbol{P}_{X', t}\) is for prompting the model with soft-perturbed text. For a given sequence input \(X = { x_{1}, \ldots, x_{t-1} }\) and a model of vocabulary size \(v\), at time step \(t\), the model generates a distribution \(\boldsymbol{P}_{X, t}\) for the next token \(x_{t}\). The final Soft-NS and Soft-NC at step \(t\) for text generation are formulated as:

\begin{equation}
\label{equ:seqential_Suff}
\small
\begin{aligned}
    \text{Soft-NS}(\mathbf{X}, x_{t}, \mathcal{R}) = \frac{\max(0, \Delta \boldsymbol{P}_{0, t} - \Delta \boldsymbol{P}_{X', t} )} {\Delta \boldsymbol{P}_{0, t}} \\
 \end{aligned}
\end{equation}
\begin{equation}
\label{equ:seqential_Comp}
\small
    \begin{aligned}
        \text{Soft-NC}(\mathbf{X}, x_{t}, \mathcal{R}) = \frac{\Delta \boldsymbol{P}_{\mathbf{X'}_{\backslash\mathcal{R}}, t}}{\Delta \boldsymbol{P}_{0, t}} \\
    \end{aligned}
\end{equation}
\noindent where \(\Delta \boldsymbol{P}_{0, t}\) is Hellinger distance between a zero input's probability distribution and full-text input's probability distribution. \({\mathbf{X'}_{\backslash\mathcal{R}}, t}\) is the case of ``if removing elements'' described in Eq.~\ref{eq:bermouli}.

\citet{cifka-liutkus-2023-black} adopt KL divergence to measure the impact on the model probability distribution when using different lengths of left-hand context. Noted by \citet{cifka-liutkus-2023-black}, this metric in their study is not related to the marginal effect of each token on the model prediction~\citep{bastings-filippova-2020-elephant, pham-etal-2022-double}, but rather to measure how much ``new information'' the token adding to the input that other tokens have not covered.\footnote{Several metrics of measuring the distance for discrete distributions exist, such as the Bhattacharyya distance, the Hellinger distance, and Kullback-Leibler divergence~\citep{lebret-collobert-2014-word, grave-etal-2014-markovian, weeds-etal-2004-characterising, o-seaghdha-copestake-2008-semantic}. We use Hellinger distance for its simplicity and symmetry property (as it is a true distance)~\citep{lebret-collobert-2014-word}. Also, its range is [0, 1], considering the \(p(\hat{y}| \mathbf{X})- p (\hat{y}|\mathcal{R})\) range [-1, 1].} 

We evaluate the \textbf{token-level faithfulness} on the dataset LongRA and the \textbf{sequence-level faithfulness} on TellMeWhy and WikiBio. Specifically, on LongRA, we evaluate Soft-NS/NC on the target word (from the word pair) for a token-level faithfulness. On TellMeWhy and WikiBio, we evaluate Soft-NS/NC for every five tokens\footnote{That is, evaluating on a stride size of five tokens. The purpose is to save computing. In our code, it is a changeable hyperparameter.} from the sequential prediction, and take the average as the sequence-level faithfulness. 

\subsection{Implementation Details}

Our method has four hyperparameters: ratio of replaced tokens \(r\) for updating, validating tolerance \(k\) and rationale size \(n\) for the stopping condition. We empirically find the faithfulness of our method is not sensitive to the hyperparameters of the following ranges: replacing ratio \(r \in ({0.1, 0.3})\), validating tolerance \(k \in ({3, 5})\), and rationale size \(n \in ({5, 8})\). A faithfulness comparison of different hyperparameters using GPT2--354M and further discussion is presented in Table~\ref{tab:hyper_sensitivity} in Section~\ref{app:hyper_sensitivity}. 
We recommend practitioners use validating tolerance \(k = 3\), replacing ratio \(r = 0.1\), and replacing ratio 0.3, which is the fastest combination, and its faithfulness is not necessarily decreased.

Our method relies on the random selection of to-be-replaced tokens.
To handle the stochasticity and the potential coverage issues, we perform three separate runs with different random seeds to obtain three importance distributions. We average distributions that stop before a maximum of 1,000 steps (loops in Algorithm~\ref{alg:pseduo_code}).

We use pre-trained generative LMs from the Huggingface library \citep{wolf-etal-2020-transformers}. We do not fine-tune any model and always use vanilla settings. Experiments are run on a single NVIDIA A100 GPU with 80GB memory (HBM2e). We use a Dell PowerEdge XE8545 machine with CUDA parallelism.

\section{Results}

\subsection{Overall Faithfulness}
\begin{figure}[!t]
\centering
    \includegraphics[trim={0 0 0 0}, width=1.\columnwidth]{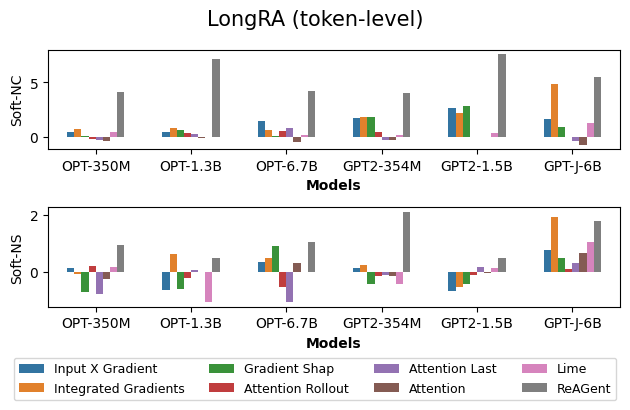}
    \caption{Token-level faithfulness on LongRA\@. Values that are close to zero indicate its faithfulness is on par with the random baseline. }\label{fig:token_level}
\end{figure}

\begin{figure}[!t]
\centering
    \includegraphics[trim={0 0 0 0}, width=1.\columnwidth]{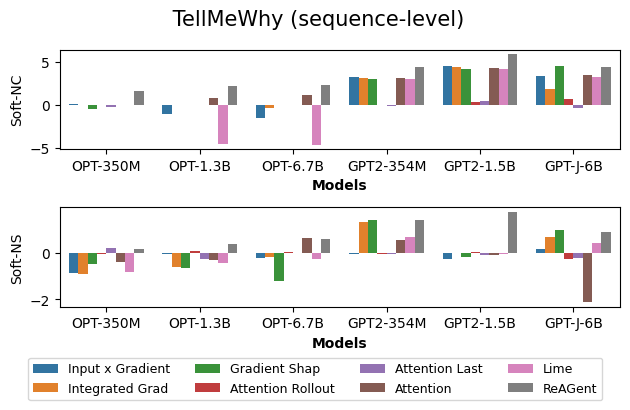}
    \includegraphics[trim={0 0cm 0 0}, width=1.\columnwidth]{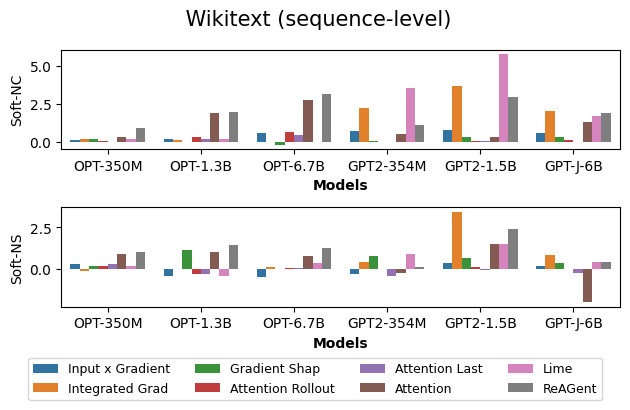}
    \caption{Sequence-level faithfulness, i.e. Soft-NS and Soft-NC, on the two datasets: WikiBio and TellMeWhy. Values that are close to zero indicate its faithfulness is on par with the random baseline.}\label{fig:sentence_level}
\end{figure}

Figure~\ref{fig:token_level} shows results on token-level faithfulness. Figure~\ref{fig:sentence_level} shows sequence-level faithfulness. All Soft-NS/NC are presented as the log of scores divided by the random baseline. Accordingly, FAs with Soft-NS/NC below zero are less faithful than the random baseline, i.e.\ unfaithful. Exhaustive results with significant tests can be found in Table~\ref{tab:full_results} in the Appendix~\ref{app:detailed_results}.

Overall, no single FA outperforms the others consistently across metrics, datasets, and models. ReAGent outperforms the other FAs in more than half of evaluation cases. For example, on LongRA (Figure~\ref{fig:token_level}), ReAGent always achieves the highest Soft-NC across models, and on WikiBio, ReAGent consistently outperforms the rest of the FAs on all OPT models. Further, ReAGent is the only FA that consistently outperforms the random baseline. Therefore, we suggest that \textbf{the faithfulness of a FA varies across generative models and generation tasks. Compared with popular FAs, our FA is more faithful as it shows faithfulness consistency, and it outperforms the other FAs across tasks and models overall.}  

Between tasks, ReAGent shows a more apparent advantage over other FAs on the token-level task, LongRA\@. ReAGent consistently outperforms the others on LongRA except for being on par with the best, integrated gradients, on Soft-NS on OPT-1.3B and GPT-J-6B. Between models, we observe that FAs tend to have comparable faithfulness performance on models from the same model family than non-family models. For example, on WikiBio, Integrated gradient always has higher Soft-NS and Soft-NC on GPT family than on OPT family. Similarly, Lime is consistently on par with the random baseline on OPT models but is the top-two highest on GPT models. We further investigate these in later sections.

\subsection{Faithfulness across Tasks and Models}
We average Soft-NS/NC over tasks (Table~\ref{tab:aggre_data_result}) and models (Table~\ref{tab:aggre_model_result}) respectively. Overall, ReAGent outperforms the other FAs in 17 column-wise comparisons out of 18 in total in the two tables. On the only one exceptional, ReAGent is on par with the corresponding best, i.e. 1.008 to 1.147 (Soft-NS, GPT-J-6B). 

\begin{table}[t]
\resizebox{\columnwidth}{!}{%
\begin{tabular}{@{}lcccccc@{}}
\toprule
 & \multicolumn{3}{c}{Soft-NC} & \multicolumn{3}{c}{Soft-NS} \\ \cmidrule(l){2-7} 
\multicolumn{1}{c}{\textbf{}} & LongRA & TellMeWhy & WikiBio & LongRA & TellMeWhy & WikiBio \\
\midrule
\multicolumn{1}{l|}{Attention} & -0.28 & 2.161 & \multicolumn{1}{l|}{1.176} & 0.099 & -0.3 & 0.302 \\
\multicolumn{1}{l|}{Attention Last} & 0.048 & 0.07 & \multicolumn{1}{l|}{0.092} & -0.222 & -0.102 & -0.151 \\
\multicolumn{1}{l|}{Attention Rollout} & 0.209 & 0.047 & \multicolumn{1}{l|}{0.211} & -0.099 & -0.085 & -0.023 \\
\multicolumn{1}{l|}{Gradient Shap} & 1.101 & 1.892 & \multicolumn{1}{l|}{0.108} & -0.116 & -0.029 & 0.51 \\
\multicolumn{1}{l|}{Input X Gradient} & 1.423 & 1.463 & \multicolumn{1}{l|}{0.49} & 0.03 & -0.22 & -0.081 \\
\multicolumn{1}{l|}{Integrated Gradients} & 1.865 & 1.536 & \multicolumn{1}{l|}{1.384} & 0.451 & 0.045 & 0.765 \\
\multicolumn{1}{l|}{Lime} & 0.412 & 0.249 & \multicolumn{1}{l|}{1.906} & -0.012 & -0.091 & 0.461 \\
\multicolumn{1}{l|}{ReAGent} & \textbf{5.402} & \textbf{4.504} & \multicolumn{1}{l|}{\textbf{1.982}} & \textbf{1.136} & \textbf{1.024} & \textbf{1.087} \\
\bottomrule
\end{tabular}%
}
\caption{Soft-NS and Soft-NC averaged over tasks. The best FA on the model (column) is highlighted in bold.}\label{tab:aggre_data_result}
\end{table}

\begin{table}[t]
\resizebox{\columnwidth}{!}{%
\begin{tabular}{@{}lllllll@{}}
\toprule
\multicolumn{1}{c}{\underline{\textbf{Soft-NC}}} & OPT-350M & OPT-1.3B & OPT-6.7B & GPT2--354M & GPT2--1.5B & GPT-J-6B \\ \midrule
Attention & 0.011 & 0.865 & 1.167 & 1.142 & 1.542 & 1.387 \\
Attention Last & -0.161 & 0.163 & 0.431 & -0.114 & 0.204 & -0.104 \\
Attention Rollout & -0.059 & 0.241 & 0.457 & 0.192 & 0.138 & -0.034 \\
Gradient Shap & -0.051 & 0.222 & -0.022 & 1.645 & 2.449 & 1.959 \\
Input x Gradient & 0.243 & -0.12 & 0.188 & 1.939 & 2.64 & 1.86 \\
Integrated Gradients & 0.323 & 0.328 & 0.129 & 2.408 & 3.442 & 2.94 \\
Lime & 0.221 & -1.431 & -1.48 & 2.269 & 3.47 & 2.086 \\
ReAGent & \textbf{2.187} & \textbf{3.753} & \textbf{5.247} & \textbf{3.202} & \textbf{5.471} & \textbf{3.916} \\ 

\midrule

\multicolumn{1}{c}{\underline{\textbf{Soft-NS}}} & OPT-350M & OPT-1.3B & OPT-6.7B & GPT2--354M & GPT2--1.5B & GPT-J-6B \\ \midrule
Attention & 0.068 & 0.233 & 0.581 & 0.039 & 0.448 & -1.168 \\
Attention Last & -0.085 & -0.173 & -0.397 & -0.21 & -0.005 & -0.079 \\
Attention Rollout & 0.089 & -0.151 & -0.214 & -0.077 & 0.006 & -0.068 \\
Gradient Shap & -0.334 & -0.035 & -0.107 & 0.586 & 0.026 & 0.593 \\
Input x Gradient & -0.149 & -0.371 & -0.133 & -0.079 & -0.181 & 0.371 \\
Integrated Gradients & -0.384 & -0.002 & 0.134 & 0.657 & 0.971 & \textbf{1.147} \\
Lime & -0.16 & -0.649 & 0.01 & 0.377 & 0.523 & 0.614 \\
ReAGent & \textbf{0.693} & \textbf{0.759} & \textbf{1.306} & \textbf{1.192} & \textbf{1.535} & 1.008 \\ 
\bottomrule

\end{tabular}
}
\caption{Soft-NS and Soft-NC averaged over models. The best FA on the model (column) is highlighted in bold. }\label{tab:aggre_model_result}
\end{table}

\begin{table}[b!]
\resizebox{\columnwidth}{!}{%
\begin{tabular}{@{}llll@{}}

\toprule

Dataset & Full Output & FA & Input \\ 

\midrule

\multirow{2}{*}{WikiBio} & \multirow{2}{*}{\parbox{2cm}{developed by \underline{\textbf{Nintendo}} for the Nintendo Entertainment System.}} & \parbox{1.2cm}{ReAGent} & \parbox{6cm}{
    \setlength{\fboxsep}{0pt}
    % gamma 8.0
    \colorbox{cyan!29.0378690509}{\strut{Super}}
    \colorbox{cyan!75.2932484768}{\strut{ Mario}}
    \colorbox{cyan!8.0643113635}{\strut{ Land}}
    \colorbox{cyan!0.3437688372}{\strut{ is}}
    \colorbox{cyan!0.0766846004}{\strut{ a}}
    \colorbox{cyan!0.0001482127}{\strut{ side}}
    \colorbox{cyan!0.0184724381}{\strut{sc}}
    \colorbox{cyan!0.1144532876}{\strut{rolling}}
    \colorbox{cyan!0.0003160420}{\strut{ platform}}
    \colorbox{cyan!1.2013254947}{\strut{ video}}
    \colorbox{cyan!0.0837805534}{\strut{ game}}
    \colorbox{cyan!11.0005185948}{\strut{developed}}
    \colorbox{cyan!0.0027606547}{\strut{ by}}
} \\ [10pt]

 &  & \parbox{1.2cm}{Lime} & \parbox{6cm}{
    \setlength{\fboxsep}{0pt}
    % gamma 8.0
    \colorbox{cyan!15.8534892875}{\strut{Super}}
    \colorbox{cyan!14.4376244284}{\strut{ Mario}}
    \colorbox{cyan!48.4124678873}{\strut{ Land}}
    \colorbox{cyan!16.5668962916}{\strut{ is}}
    \colorbox{cyan!1.2736685089}{\strut{ a}}
    \colorbox{cyan!31.3983287669}{\strut{ side}}
    \colorbox{cyan!61.4902888256}{\strut{sc}}
    \colorbox{cyan!40.3366185699}{\strut{rolling}}
    \colorbox{cyan!41.4951703668}{\strut{ platform}}
    \colorbox{cyan!25.9773520648}{\strut{ video}}
    \colorbox{cyan!23.3343624499}{\strut{ game}}
    \colorbox{cyan!24.9878323250}{\strut{developed}}
    \colorbox{cyan!52.3504528296}{\strut{ by}}
} \\ [20pt]

\midrule
 
\multirow{2}{*}{TellMeWhy} & \multirow{2}{*}{\parbox{2cm}{He \underline{\textbf{went}} to see his old college.}} & \parbox{1.2cm}{ReAGent} & \parbox{6cm}{
    \setlength{\fboxsep}{0pt}

    % gamma 8.0
    \colorbox{cyan!0.1098677230}{\strut{Jay}}
    \colorbox{cyan!0.0054770008}{\strut{ took}}
    \colorbox{cyan!0.0017670759}{\strut{ a}}
    \colorbox{cyan!7.3739967035}{\strut{ trip}}
    \colorbox{cyan!0.0120127702}{\strut{ to}}
    \colorbox{cyan!0.0661700770}{\strut{ his}}
    \colorbox{cyan!0.0198750700}{\strut{ old}}
    \colorbox{cyan!0.0596416326}{\strut{ college}}
    \colorbox{cyan!0.0668745035}{\strut{}}
    \colorbox{cyan!3.0514559673}{\strut{ Jay}}
    \colorbox{cyan!0.0001801143}{\strut{ is}}
    \colorbox{cyan!0.2862125799}{\strut{ an}}
    \colorbox{cyan!7.1505334480}{\strut{ alumni}}
    \colorbox{cyan!0.1978298635}{\strut{}}
    \colorbox{cyan!0.0044185178}{\strut{ He}}
    \colorbox{cyan!0.0103834291}{\strut{ visited}}
    \colorbox{cyan!0.8570611903}{\strut{ his}}
    \colorbox{cyan!0.0212062075}{\strut{ friends}}
    \colorbox{cyan!0.0001713511}{\strut{ He}}
    \colorbox{cyan!95.1035709771}{\strut{ went}}
    \colorbox{cyan!0.0019330349}{\strut{ and}}
    \colorbox{cyan!0.0062024045}{\strut{ got}}
    \colorbox{cyan!0.0017818837}{\strut{ drunk}}
    \colorbox{cyan!0.0220165374}{\strut{}}
    \colorbox{cyan!0.0002646773}{\strut{ He}}
    \colorbox{cyan!0.1029187666}{\strut{ had}}
    \colorbox{cyan!2.5034189441}{\strut{ a}}
    \colorbox{cyan!0.0072111149}{\strut{ good}}
    \colorbox{cyan!0.0286259103}{\strut{ time}}
    \colorbox{cyan!2.0727392716}{\strut{}}
    \colorbox{cyan!0.0083288485}{\strut{ Why}}
    \colorbox{cyan!3.9135548934}{\strut{ did}}
    \colorbox{cyan!2.4005896355}{\strut{ He}}
    \colorbox{cyan!94.5587476139}{\strut{ go}}
    \colorbox{cyan!0.8612037974}{\strut{?}}
    \colorbox{cyan!96.0741101662}{\strut{He}}

} \\ [20pt]

 &  & \parbox{1.2cm}{Lime} & \parbox{6cm}{
    \setlength{\fboxsep}{0pt}

    % gamma 9.0
    \colorbox{cyan!15.7219068381}{\strut{Jay}}
    \colorbox{cyan!7.9846428348}{\strut{ took}}
    \colorbox{cyan!21.0971093551}{\strut{ a}}
    \colorbox{cyan!19.3540861652}{\strut{ trip}}
    \colorbox{cyan!18.0788093521}{\strut{ to}}
    \colorbox{cyan!33.6546585020}{\strut{ his}}
    \colorbox{cyan!6.6803600658}{\strut{ old}}
    \colorbox{cyan!14.4203020277}{\strut{ college}}
    \colorbox{cyan!12.0794326674}{\strut{}}
    \colorbox{cyan!26.1818564893}{\strut{ Jay}}
    \colorbox{cyan!1.2125873573}{\strut{ is}}
    \colorbox{cyan!12.0178844872}{\strut{ an}}
    \colorbox{cyan!19.1710664820}{\strut{ alumni}}
    \colorbox{cyan!4.3137080790}{\strut{}}
    \colorbox{cyan!10.2775295599}{\strut{ He}}
    \colorbox{cyan!4.3392829686}{\strut{ visited}}
    \colorbox{cyan!27.7316410476}{\strut{ his}}
    \colorbox{cyan!28.2482682400}{\strut{ friends}}
    \colorbox{cyan!23.0965718989}{\strut{}}
    \colorbox{cyan!7.6220993927}{\strut{ He}}
    \colorbox{cyan!23.8130554762}{\strut{ went}}
    \colorbox{cyan!30.4149377785}{\strut{ and}}
    \colorbox{cyan!39.4311733248}{\strut{ got}}
    \colorbox{cyan!29.1249370978}{\strut{ drunk}}
    \colorbox{cyan!1.2537159760}{\strut{}}
    \colorbox{cyan!2.3413398771}{\strut{ He}}
    \colorbox{cyan!12.0261144022}{\strut{ had}}
    \colorbox{cyan!31.3525708938}{\strut{ a}}
    \colorbox{cyan!16.8977284087}{\strut{ good}}
    \colorbox{cyan!7.9367548558}{\strut{ time}}
    \colorbox{cyan!18.7815050565}{\strut{}}
    \colorbox{cyan!4.8695337588}{\strut{ Why}}
    \colorbox{cyan!16.8348111951}{\strut{ did}}
    \colorbox{cyan!21.4492322422}{\strut{ He}}
    \colorbox{cyan!60.8375804118}{\strut{ go}}
    \colorbox{cyan!13.5888224363}{\strut{?}}
    \colorbox{cyan!38.8422420773}{\strut{He}}
    
} \\

\bottomrule
\end{tabular}%
}

\caption{Importance distribution is given by ReAGent and Lime, for Model GPT2--1.5B.}\label{tab:seq_heatmap_result}
\end{table}

\begin{figure}[ht!]
\centering
    \includegraphics[trim={0 0 0 0}, width=.99\columnwidth, height=0.44\linewidth]{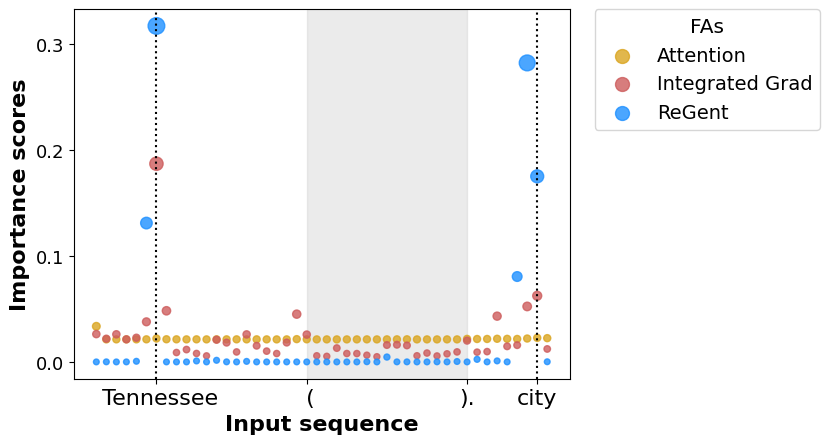}
    \caption{Importance distribution over the input: \textit{``As soon as I arrived in Tennessee, I checked into my hotel, and watched a movie before falling asleep. (I had a great call with my husband, although I wish it were longer). I was staying in my favorite city, ''}. The sentence in \colorbox{lightgray}{()} is the distractor. The model predicts \textit{``Nashville''} regardless of whether the input includes the distractor or not. %Values in () in the legend indicate the Soft-NS and Soft-NC for this input.
    }\label{fig:distribution_example}
\end{figure}

\paragraph{Tasks} As shown in Figure~\ref{tab:aggre_data_result}, ReAGent is the most faithful FA in terms of both Soft-NS and Soft-NC across task-wise comparison.
Particularly, for the token-level task, LongRA, ReAGent overwhelmingly outperforms the second-highest, i.e. 5.402 to 1.865 (Integrated Gradients). In task LongRA, intuitively, the model takes more hints from the indicator token in the input, e.g. ``Tennessee'' to ``Nashville''. Between TellyMeWhy and WikiBio, ReAGent exhibits greater advantages on TellyMeWhy, where the prompts provide richer contextual information. We conduct a qualitative analysis of the importance distributions by each FA and observe that ReAGent more effectively captures the hint token within the context of the input prompt, examples from GPT2--1.5B are shown in Figure~\ref{fig:distribution_example} and Table~\ref{tab:seq_heatmap_result}.

Figure~\ref{fig:distribution_example} shows the importance distribution given by the best (ReAGent), the second-best (Integrated gradients), and the worst (Attention) FAs on LongRA\@. For this example, ReAGent is able to better catch the importance of \textit{``Tennessee''} and \textit{``city''} than Integrated Gradients and Attention. Although Integrated gradients, the second-best FA, recognizes the importance of \textit{``Tennessee''}, it fails to ignore the contribution of the distractor sentence as ReAGent does. Likely, we see the WikiBio example in Table~\ref{tab:seq_heatmap_result}, ReAgent more emphasizes the importance of ``Super Mario'' than Lime when predicting ``Nintendo''.

We conclude that \textbf{for tasks provided with contextualized information and expecting specific predictions, ReAGent is more faithful than the other FAs. It is capable to pick up the link between the hint in the input and the predictions that the model utilizes to make the specific prediction.}

\paragraph{Models} As shown in Figure~\ref{tab:aggre_data_result}, between OPT and GPT family, ReAGent shows greater advantages on OPT models than on GPT models, i.e.\ ReAGent consistently outperforms other FAs on OPT models on both metrics while only achieving the top-two consistently on GPT models. This indicates the possible difference in inner operation between model families and the similarity within model families.

\subsection{Evaluation on Long-Range Agreement with Greedy Search}

Following~\citet{vafa-etal-2021-rationales}, we conduct a Long-Range Agreement evaluation of rationales given by each FA and the greedy search method (i.e.\ greedy rationalization~\citep{vafa-etal-2021-rationales}).%, which is different from the faithfulness evaluation of the overall importance distribution in the paper. 
The greedy search aims to discover the smallest subset of input tokens that would predict the same output as the full sequence~\citep{vafa-etal-2021-rationales}. Therefore, rationales given by the greedy search are in various lengths, i.e.\ each input has its own rationales in different lengths. We reproduce their experiment and apply the input-specific rationale length by greedy search to extract the rationale for each FA, and then evaluate the Ante ratio and No.D ratio of rationales. We acknowledge the potential bias that the setup of the rationale length is biased towards the greedy search due to the limitation of the greedy search itself. Specifically, greedy search only provides binary rationales, i.e.\ a token is a rationale or not, rather than an importance distribution. Further, for a given output, two selected rationales by greedy search will be taken as equally important. We only experiment with GPT2--354M as greedy search requires further fine-tuning.

Table~\ref{tab:task1_results} shows the Ante ratio and No.D ratio of each FA when using varying lengths of rationales by greedy search, on dataset LongRA\@. For both metrics, the higher, the better the rationales are, according to the faithfulness definition (the smallest subset for predicting the same output as the full sequence) by \citet{vafa-etal-2021-rationales}. As shown in the table, although the setup of rationale length favours greedy search by nature, Attention Rollout outperforms greedy search on No.D and Gradient Shap is on par with Greedy Search on Ante.

\begin{table}[]
\centering
\small
\begin{tabular}{lcc}
    \toprule
    & Ante Ratio & No.D Ratio \\
    \midrule 
    Gradient Shap  & \textbf{1} & 0.592 \\
    Input x Gradient  & 0.471 & 0.471 \\
    Integrated Gradients  & 0.987 & 0.503 \\
    Attention Rollout & 0.357 & \textbf{0.930} \\
    Attention Last  & 0.898 & 0.745 \\
    Attention & 0.936 & 0.707 \\
    Greedy Search & \textbf{1} & 0.667 \\
    ReAGent & 0.930 & 0.710 \\ % configure: top3_replace0.1_max3000_batch5
    \bottomrule
\end{tabular}
\caption{FAs faithfulness on long-range agreement with templates analogies. ``Ratio'' refers to the approximation ratio of each method's rationale length to the exhaustive search minimum. ``Ante''  refers to the percentage of rationales that contain the true antecedent. ``No D'' refers to the percentage of rationales that do not contain any tokens from the distractor.}\label{tab:task1_results}
\end{table}

\subsection{Impact of Hyper-Parameters}\label{app:hyper_sensitivity}

All of our experiments do not involve training or fine-tuning any language models.
Gradient-based FAs and Lime are built upon Inseq library \citep{sarti-etal-2023-inseq}. 
All reported results of ReAGent in the main body are with a replacing ratio of 30\%, max step of 3000, three runs, and top five unimportant tokens for stopping conditions. 

To examine the sensitivity of hyper-parameters of ReAGent, we experiment on GPT2--354M with different settings of these hyperparameters. The result shows that our method is always more faithful than the random baseline (above zero), as shown in Table~\ref{tab:hyper_sensitivity}.

\begin{table}[ht!]
\centering
\resizebox{\columnwidth}{!}{%
\begin{tabular}{@{}cllll@{}}
\toprule
Replacing Ratio \textbf{r} & Max Step \textbf{m} & Runs & Soft-NS & Soft-NC \\
\midrule
0.1 & 5000 & 5 & 1.910 & 1.623 \\
0.1 & 5000 & 3 & 2.089 & 3.009 \\
0.1 & 3000 & 5 & 1.752 & 2.170 \\
0.3 & 3000 & 3 & 1.752 & 2.170 \\
0.3 & 3000 & 5 & 2.09 & 4.009 \\
0.3 & 3000 & 3 & 2.406 & 1.481 \\
0.3 & 5000 & 3 & 2.039 & 1.639 \\
\bottomrule
\end{tabular}%
}
\caption{Soft-NS and Soft-NC of ReAGent on LongRA, with different hyperparameters. }\label{tab:hyper_sensitivity}
\end{table}

\section{Conclusion}
We proposed ReAGent, a model-agnostic feature attribution method for text generation tasks. Our method does not require fine-tuning, any model modification, or accessing the internal weights. We conducted extensive experiments across six decoder-only models of two model families in different sizes. The results demonstrate that our method is consistently more faithful than other commonly-used FAs. 

In future work, we plan to evaluate our method on different generative models, e.g.\ encoder-decoder and diffusion models, and different text generation tasks, e.g.\ translation and summarization.  

%%%%%%%%%%%%%%%%%%%%%% APPENDIX BEFORE REFERENCE FOR AAAI

%%%%%%%%%%%%%%%%%%%%%% Acknowledgments BEFORE REFERENCE FOR AAAI
\appendix

\section{Replacement tokens}\label{app:alternative_find_token_replace}

\begin{table}[]
\centering
\resizebox{\columnwidth}{!}{%
\begin{tabular}{@{}llllll@{}}
\toprule
Model & Token replacing method & Steps & Steps std & Soft-NS & Soft-NC \\ \midrule
\multirow{3}{*}{GPT2 354M} & RoBERTa & \textbf{244} & 633 & \textbf{2.358} & 2.269 \\
 & Random & 4041 & 1645 & 1.95 & \textbf{5.578} \\
 & POS & 2154 & 1142 & 1.35 & 2.028 \\ \midrule
\multirow{3}{*}{GPT2 1.5B} & RoBERTa & \textbf{239} & 701 & 1.215 & 5.105 \\
 & Random & 4379 & 1345 & \textbf{1.344} & \textbf{8.173} \\
 & POS & 2434 & 928 & 0.039 & 3.419 \\ \midrule
\multirow{3}{*}{GPT-J 6B} & RoBERTa & \textbf{364} & 972 & 1.593 & 2.739 \\
 & Random & 3126 & 2328 & 0.943 & \textbf{6.908} \\
 & POS & 2105 & 1285 & \textbf{1.758} & 3.217 \\ \midrule
\multirow{3}{*}{OPT 350M} & RoBERTa & \textbf{108} & 106 & 1.483 & 3.141 \\
 & Random & 3776 & 1686 & 1.527 & \textbf{3.947} \\
 & POS & 1554 & 1132 & 2.055 & 1.723 \\ \midrule
\multirow{3}{*}{OPT 1.3B} & RoBERTa & \textbf{265} & 379 & 0.812 & 4.657 \\
 & Random & 4291 & 1362 & \textbf{1.141} & \textbf{7.511} \\
 & POS & 2642 & 827 & 0.865 & 3.308 \\ \midrule
\multirow{3}{*}{OPT 6.7B} & RoBERTa & \textbf{259} & 391 & 0.876 & 7.486 \\
 & Random & 4878 & 522 & 0.884 & \textbf{9.426} \\
 & POS & 2810 & 534 & \textbf{1.825} & 4.725 \\ \bottomrule
\end{tabular}%
}
\caption{Soft-NS and Soft-NC results of different token replacing methods. As other Soft-NS and Soft-NC results presented in the paper, results are presented as the log of scores divided by the random baseline. Maximum steps are set to 3000.}\label{tab:alternative_method}
\end{table}

\begin{figure}[ht!]
\centering
    \includegraphics[trim={0 1.22cm 0 0}, width=.79\columnwidth, height=0.44\linewidth]{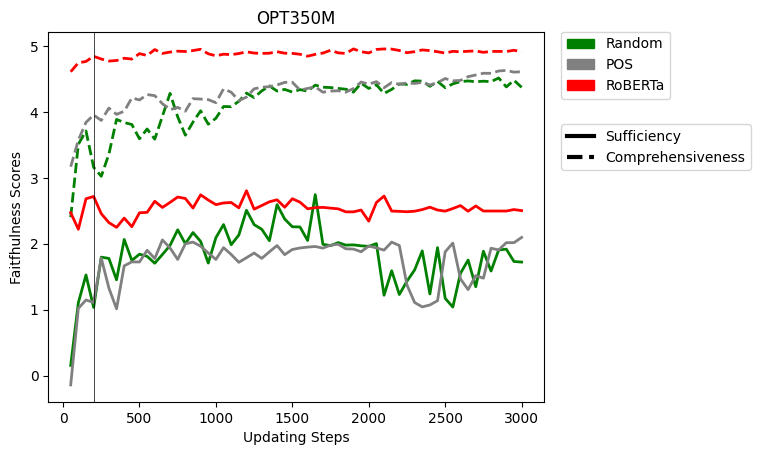}
    \caption{Sufficiency and Comprehensiveness scores on different updating steps.
    }\label{fig:converage}
\end{figure}

ReAGent measures the predictive likelihood changes after top-k important tokens are perturbed. The perturbation operation replaces these top-k tokens with RoBERTa predicted tokens. We also experimented with replacing them with random tokens from the vocabulary (\(\hat{x}_i = U(\mathcal{V})\), where \(\mathcal{V}\) is the vocabulary and \(i \in [0, t]\)), noted as ``Random'', and replacing them with random tokens of the same POS, noted as ``POS''.

When replacing top-k with random tokens or tokens of the same POS, we found it often takes thousands of steps to meet the stopping condition, as shown by the average ``Steps'' in Table~\ref{tab:alternative_method}.\footnote{In experiments, we set up a maximum updating steps of 5000.} Figure~\ref{fig:converage} shows the faithfulness scores at different steps (evaluating per 50 steps). We see that using RoBERTa predicted tokens to replace, both sufficiency and comprehensiveness coverage much earlier than Random and POS replacements, as shown in the black vertical line in Figure~\ref{fig:converage}. On the other hand, replacing top-k important tokens with random tokens or tokens of the same POS does not bring consistently higher faithfulness.

\section{Full Faithfulness Results}\label{app:detailed_results}

\begin{table*}[ht!]
\centering
\resizebox{0.8\textwidth}{!}{%
\begin{tabular}{@{}llcccccc@{}}
\toprule
 &  & \multicolumn{2}{c}{LongRA} & \multicolumn{2}{c}{TellMeWhy} & \multicolumn{2}{c}{Wikitext} \\ \midrule
Model & FAs & Soft-NS & Soft-NC & Soft-NS & Soft-NC & Soft-NS & Soft-NC \\ \midrule
GPT2--354M & Input x Gradient & 0.145 & 1.793 & -0.062 & 3.3 & -0.32* & 0.725 \\
GPT2--354M & Integrated Gradients & 0.266 & 1.827 & 1.293* & 3.152 & 0.411* & 2.244 \\
GPT2--354M & Gradient Shap & -0.419 & 1.867 & 1.405 & 3.046 & 0.773* & 0.023* \\
GPT2--354M & Attention Rollout & -0.125* & 0.487* & -0.07* & 0.069* & -0.035* & 0.02* \\
GPT2--354M & Attention Last & -0.098* & -0.281* & -0.054* & -0.036* & -0.477* & -0.027* \\
GPT2--354M & Attention & -0.137* & -0.226* & 0.53* & 3.141 & -0.275* & 0.511 \\
GPT2--354M & Lime & -0.404 & 0.168 & 0.67* & 3.057 & 0.865 & 3.581 \\
GPT2--354M & ReAGent & 2.09 & 4.009 & 1.404 & 4.483 & 0.081 & 1.113 \\
GPT2--1.5B & Input x Gradient & -0.648 & 2.64 & -0.256* & 4.529 & 0.363* & 0.752* \\
GPT2--1.5B & Integrated Gradients & -0.513* & 2.165 & -0.011* & 4.443 & 3.437 & 3.72 \\
GPT2--1.5B & Gradient Shap & -0.398* & 2.823 & -0.183* & 4.186 & 0.66* & 0.339* \\
GPT2--1.5B & Attention Rollout & -0.08 & -0.008* & 0.015* & 0.348* & 0.083* & 0.073* \\
GPT2--1.5B & Attention Last & 0.17* & 0.052* & -0.093* & 0.533* & -0.091* & 0.029* \\
GPT2--1.5B & Attention & -0.034* & 0.046* & -0.099* & 4.285 & 1.479* & 0.295* \\
GPT2--1.5B & Lime & 0.131* & 0.366 & -0.041* & 4.257 & 1.477 & 5.787 \\
GPT2--1.5B & ReAGent & 0.492 & 7.556 & 1.735 & 5.927 & 2.378 & 2.931 \\
GPT-J-6B & Input x Gradient & 0.784 & 1.652 & 0.173* & 3.366 & 0.157* & 0.56 \\
GPT-J-6B & Integrated Gradients & 1.923 & 4.885 & 0.68* & 1.89* & 0.838 & 2.045 \\
GPT-J-6B & Gradient Shap & 0.497 & 0.948 & 0.945 & 4.592 & 0.337 & 0.337 \\
GPT-J-6B & Attention Rollout & 0.114* & 0.015* & -0.27* & 0.694* & -0.039* & 0.128* \\
GPT-J-6B & Attention Last & 0.303* & -0.35* & -0.225* & -0.337* & -0.292* & -0.014* \\
GPT-J-6B & Attention & 0.68 & -0.671* & -2.128 & 3.544 & -2.056 & 1.288 \\
GPT-J-6B & Lime & 1.04 & 1.264 & 0.398 & 3.278 & 0.404 & 1.717 \\
GPT-J-6B & ReAGent & 1.764 & 5.477 & 0.875 & 4.386 & 0.384 & 1.885 \\
OPT-350M & Input x Gradient & 0.152* & 0.49 & -0.879* & 0.11 & 0.281* & 0.131* \\
OPT-350M & Integrated Gradients & -0.067* & 0.79 & -0.915* & 0.019* & -0.17* & 0.161* \\
OPT-350M & Gradient Shap & -0.681* & 0.151 & -0.481 & -0.474 & 0.16* & 0.169* \\
OPT-350M & Attention Rollout & 0.211* & -0.192* & -0.076* & -0.015* & 0.134* & 0.032* \\
OPT-350M & Attention Last & -0.739* & -0.295* & 0.209 & -0.152* & 0.275* & -0.037* \\
OPT-350M & Attention & -0.248* & -0.313* & -0.412* & 0.033 & 0.864 & 0.312 \\
OPT-350M & Lime & 0.197* & 0.491 & -0.834* & -0.029* & 0.156* & 0.202* \\
OPT-350M & ReAGent & 0.932 & 4.074 & 0.147 & 1.6 & 0.999 & 0.887 \\
OPT-1.3B & Input x Gradient & -0.617 & 0.497 & -0.065* & -1.049* & -0.43* & 0.193 \\
OPT-1.3B & Integrated Gradients & 0.618 & 0.858 & -0.603* & 0* & -0.022* & 0.127* \\
OPT-1.3B & Gradient Shap & -0.588* & 0.696 & -0.647* & 0.002* & 1.132 & -0.034* \\
OPT-1.3B & Attention Rollout & -0.193* & 0.353 & 0.064* & 0.035 & -0.324* & 0.333 \\
OPT-1.3B & Attention Last & 0.082* & 0.321* & -0.256* & -0.008* & -0.345* & 0.176 \\
OPT-1.3B & Attention & 0.008* & -0.092* & -0.331* & 0.8 & 1.024 & 1.888 \\
OPT-1.3B & Lime & -1.032* & 0.014 & -0.452* & -4.489* & -0.462* & 0.183* \\
OPT-1.3B & ReAGent & 0.497 & 7.104 & 0.365 & 2.208 & 1.413 & 1.947 \\
OPT-6.7B & Input x Gradient & 0.365* & 1.466 & -0.229 & -1.478 & -0.536 & 0.576 \\
OPT-6.7B & Integrated Gradients & 0.481* & 0.667 & -0.175* & -0.289* & 0.096* & 0.011* \\
OPT-6.7B & Gradient Shap & 0.896 & 0.122* & -1.214 & -0.002* & -0.003* & -0.186* \\
OPT-6.7B & Attention Rollout & -0.523* & 0.6 & 0.009* & 0.02 & 0.045* & 0.681 \\
OPT-6.7B & Attention Last & -1.048 & 0.839 & -0.022* & 0.074 & 0.024* & 0.424 \\
OPT-6.7B & Attention & 0.327* & -0.426* & 0.641 & 1.166 & 0.775 & 2.76 \\
OPT-6.7B & Lime & -0.006* & 0.17* & -0.288* & -4.578* & 0.323* & -0.033* \\
OPT-6.7B & ReAGent & 1.039 & 4.188 & 0.574 & 2.36 & 1.264 & 3.13 \\ \bottomrule
\end{tabular}%
}
\caption{Full Soft-NS and Soft-NC results of different FAs on different Models. For consistency and comparison across data and models, results are presented as the log of scores divided by the random baseline. Asterisk (*) denotes the case where its p-value is greater than .05 in the Wilcoxon signed-rank Test, i.e.\ the score, Soft-NS or Soft-NC, is not significantly different from the random baseline.}\label{tab:full_results}
\end{table*}

\newpage

\section*{Acknowledgments}
We thank Professor Nikolaos Aletras and Dr George Chrysostomou for their helpful comments, thoughts, and discussions. We acknowledge IT Services at The University of Sheffield for the provision of services for High Performance Computing.

\bibliography{ms}

\end{document}